# Deep Active Learning for Remote Sensing Object Detection

Zhenshen Qu[1], Jingda Du[1]*, Yong Cao[2], Qiuyu Guan[1], Pengbo Zhao[1]
[1] Harbin Institute of Technology, Harbin, China
[2] Beijing Orient Institute of Measurement and Test, Beijing, China
e-mail: miraland@hit.edu.cn, fine0430@outlook.com, reidcao@126.com, frenchfries416600@163.com, 15546167605@163.com

*Abstract*—Recently, CNN object detectors have achieved high accuracy on remote sensing images but require huge labor and time costs on annotation. In this paper, we propose a new uncertainty-based active learning which can select images with more information for annotation and detector can still reach high performance with a fraction of the training images. Our method not only analyzes objects' classification uncertainty to find least confident objects but also considers their regression uncertainty to declare outliers. Besides, we bring out two extra weights to overcome two difficulties in remote sensing datasets, class-imbalance and difference in images' objects amount. We experiment our active learning algorithm on DOTA dataset with CenterNet as object detector. We achieve same-level performance as full supervision with only half images. We even override full supervision with 55% images and augmented weights on least confident images.

## I. INTRODUCTION

Deep learning has achieved high performance on many object detection tasks like remote sensing detection, pedestrian detection, face detection [1, 2, 3, 4]. To improve detector's accuracy, people make great efforts to propose new network structures and deepen backbone with numerous parameters and complex inner relations [5, 6, 7, 8]. However, a detector's performance is not only related with its architecture but also about training dataset's scale and quality. We usually have excellent performance on high quality datasets like [9, 10] and relatively poor performance on self-made datasets. Therefore, comprehensive and diverse datasets are necessary in proceeding of training a detector to reach considerable accuracy. With development of remote sensing techniques, high quality images are easily available and more analysis can be realized. However, some analysis methods like object detection, semantic segmentation are supervised by annotations which require high time and labor costs. We cannot annotate all available samples and have to make a choice.

In object detection datasets, we usually annotate an object with category attribute and a bounding box and annotation costs are decided by objects' amount. While objects in remote sensing images make it impossible to establish a comprehensive dataset. Usually, we have to randomly choose a limited subset for annotation and detectors' train. Unless the subset is large enough, this solution cannot guarantee detector's accuracy and generalization ability. Active learning methods can select meaningful samples according to detector's inference on unlabeled sample pool. An effective algorithm could select samples with more useful information for object detectors' training and ignore outliers which may decrease detector's performance and waste annotation cost. For normal object detection datasets, active learning algorithms usually define sum of objects' uncertainty as the image's uncertainty. However, number of objects in remote sensing image ranges widely and each category's object amount also ranges widely. These algorithms may only select some particular images and cause class-imbalance in training iterations. As a result, some categories have excellent performance while others get poor performance by few training data. Also, remote sensing objects have many kinds of sizes and even in one category their sizes range widely. Limited by annotation cost, we should make a decision on what kind of object should be labeled. Under these premises, what kind of images and objects should be labeled first?

In this paper, we propose a new weighted classification-regression uncertainty (WCR) active learning algorithm for remote sensing object detection. Our algorithm use regression uncertainty to select images having popular sizes and adapt classification uncertainty to select images having more information for detector's training. Regression uncertainty strategy intends to annotate objects having popular sizes first. With limited labeled images, we choose to perform well on majority first. Remote sensing datasets have class-imbalance and number of objects in each image ranges widely. Therefore, we propose two weights to overcome the difficulties. Our algorithm infers all the unlabeled images' detection which results to selecting those with more potential for training to increase detection accuracy. Then these images are labeled and utilized for retraining the detector. Given an image, we firstly use detector to predict its objects and then use our active learning method to calculate their uncertainty scores. After calculating all images' uncertainty scores, we rank them and select those images with high uncertainty for labeling which brings more useful information to object detector. This procedure can be performed in several iterations. Our method is suitable to object detectors which infer a bounding box and one classification probability. We choose the object detector, CenterNet [11] for experiments which directly predicts objects' classification and bounding box's center coordinates, width and height. We experiment our algorithm on DOTA dataset with CenterNet and results are displayed in Section 4.

## II. RELATED WORK

### A. Object detection using CNNs

Since [12] was proposed, object detectors based on deep learning keep improving their high accuracy. After R-CNN, some one-stage and two-stage object detectors were brought out. [13] and [14] are two well-known two-stage object detectors. They use deep neural networks to extract features and predict objects' category. They adopt anchor strategy to infer bounding boxes' position which requires high memory and computation cost. [15] is another two-stage detector which reaches high accuracy on extremely plenty categories' object detection. Two-stage object detectors achieve high accuracy with low inference speed. While one-stage object detectors have great advantage on inference speed with some sacrifice on accuracy and some of them can even realize real-time object detection. [16] is a famous one-stage object detection series which has very fast inference speed with considerable accuracy. To reach as high performance as two-stage detectors, [17] was proposed and solved the problem that easy training objects block detector's advanced learning. Besides, key-point based object detectors like [18], [11] come out in recent times. They regard objects as key-points and predict key-points to estimate objects' location information. They are more suitable to detect irregular or oriented objects. Actually, all three kinds of methods have been used on remote sensing object detection and we select a key-point based method, CenterNet for our experiment.

### B. Active learning in classification

Active learning has been widely used in image classification field and it selects most helpful unlabeled images for annotation. [19] designs an active learning algorithm for cost-sensitive multiclass classification with problems where different errors have different costs. Their algorithm, COAL, makes predictions by regressing to each label's cost and predicting the smallest. [20] proposes a novel Gaussian process classifier model with multiple annotators for multi-class visual recognition. And a generalized EM-EP algorithm is derived by them to estimate the parameters and approximate Bayesian inference. The current trend is to improve classification performance using deeper and deeper neural networks, the size of the training dataset must grow at the same time. [21] presents an active learning strategy based on query by committee and dropout technique to train a Convolutional Neural Network.

### C. Active learning in object detection

Although active learning has a long history, there are not too much active leaning algorithms for deep object detection. In 2006, [22] proposed an active learning algorithm for pedestrian detection on videos shot by a camera fixed on a vehicle. The object detection method is based on AdaBoost and unlabeled instances' selection is achieved by hand-tuned thresholds of detections. [23] brought active learning into satellite images' object detection. They used sliding window and SVM as classifier to detect objects. The combination of active learning for object detection with crowd sourcing is presented by [24] in 2014. A part-based detector for SVM classifiers in combination with hashing is proposed for use in large-scale settings. Active learning is realized by selecting the most uncertain instances to be annotated. [25] selected SSD as detector and regarded it as a white box. They develop a novel active learning method which poses the layered architecture used in object detection as a QBC paradigm to choose the set of images to be queried. [26] proposed an active learning algorithm which is suitable to detector based on convolutional neural network. They proposed a novel image-level scoring process to rank unlabeled images for automatic selection. This active learning algorithm can be applied to videos and consecutive image sets.

In this paper, we propose an active learning algorithm to select images for annotation by unlabeled images' uncertainty and labeled images' distribution. Our method is suitable to most object detectors and can save extensive labor and time costs.

## III. METHODOLOGY

As an active learning algorithm for remote sensing object detection, we will first introduce active learning with classification, then introduce active learning with regression and at last explain how to implement the two algorithms jointly on remote sensing object detection. Our active learning method is pool-based which means we just select images from limited unlabeled images pool. Here we use CenterNet for experiment and it directly infers objects' classification and location. The remote sensing dataset we use is DOTA and we prove our active learning algorithms on it.

### A. Weighted Classification Uncertainty Sampling Active Learning

To save high annotation cost for remote sensing images, we propose a new classification uncertainty sampling active learning algorithm. Different from some famous object detection datasets like [9, 27], images in remote sensing datasets are huge and have numerous objects in each. Moreover, some objects like small-vehicle have small size and are densely distributed on images. Traditional uncertainty sampling algorithms may not work well enough because they may focus on images with more objects. As a result, some classes have more and more objects during active learning training proceeding while other classes are almost ignored. When we train detector with selected images, only a few classes have better performance while other classes are even worse than before.

To achieve better performance with less annotation cost on remote sensing datasets, we propose a new active learning algorithm which selects useful images and decreases the influence from class-imbalance. When we train an object detector, the amount of objects in each class is the main reason for class-imbalance and it should be taken into account in active learning algorithm design. Here we calculate each class's ratio as, $n_i$ is each class's objects' amount:

$$W_i^1 = log_{10}^{n_i}$$

Amount only is not enough for design because one image is the smallest unit in selection and we should also estimate objects' distribution in single image. Therefore, average number of objects for each class in one image is another important parameter for images' selection. Here we calculate this ratio as:

$$\begin{cases} x_i = \dfrac{n_i}{p_i} \\ sum = \displaystyle\sum_{i=1}^{C} x_i \\ W_i^2 = (sum + C)/(x_i + 1) \end{cases}$$

$p_i$ denotes amount of images have one or more class $i$'s objects.

We select Least Confident strategy as uncertainty sampling algorithms and gather it with ratios above to select helpful images. For every image, its uncertainty can be calculated as below, algorithm[1]:

$$U_c = \sum W_i^1 * W_i^2 * (1 - P)$$

Then we rank images with their uncertainty and select images with high uncertainty which also means their contribution to detectors. This active learning method only need detectors to output pseudo probabilities for objects and can be implemented with most object detector.

### B. Regression Uncertainty Sampling Active Learning

The active learning algorithm above only considers classification uncertainty but object detectors also estimate objects' location. Therefore, we make great efforts to design active learning algorithm based on regression uncertainty and our method needs detector to estimate objects' width and height. For each bounding box, it has a long side: $L = \max(w, h)$ and a short side: $S = \min(w, h)$. We use these two parameters to calculate labeled objects' distribution probability density and use it to estimate bounding boxes' uncertainty. The design of regression uncertainty aims to find unlabeled images with more useful information. When we train a network with limited data amount, it's better to focus on the main object parts. Because with same extent improvement of precision, majority can bring larger increasement on mAP than minority. Therefore, with same classification uncertainty, we tend to select objects which have bigger probability density. And the regression uncertainty is positively related with bounding boxes' distribution probability density.

Here we select Gaussian Mixture Model (GMM) to estimate bounding boxes' distribution probability density. Firstly, we calculate each bounding box's log probability as $L$. Then we clip $L$ at -99, $L_b = \min(-99, L)$. Finally, we use $L_b$ to calculate regression uncertainty $U_r$ as algorithm[2]:

$$\begin{cases} U_r = 0.05 * (L_b + 10) + 0.5, L_b \geq -10 \\ U_r = 0.5 * \dfrac{L_b + 100}{90}, L_b < -10 \end{cases}$$

With this uncertainty formulation, we can select majority and ignore outliers which decrease network's performance while training. Also, this formulation can also be used to labeled images' contribution to network's training. Images have higher regression uncertainty should have more backpropagation weights or more training times and this implementation can improve the object detector's performance without extra annotation cost. Besides, when we use active learning to chase new images for annotation, we have to select some images randomly for initialization. However, we cannot guarantee all initial images are not outliers. When we meet the training bottleneck that only increasing labeled images' amount have little influence on performance, we can also use the regression uncertainty algorithm to delete outliers to achieve higher performance. Especially in the remote sensing object detection field, all images are huge in size and we must crop them into pieces for training. Meanwhile, there many regions which have few objects exist in the images. Therefore, it's necessary to select useful images for annotation and clear outliers from training dataset.

### C. WCR Active Learning for Remote Sensing Object Detection

Overall, we design our weighted classification-regression (WCR) active learning algorithm with algorithm[1] and algorithm[2]. The only demand for object detectors is that they should predict objects' width and height directly or indirectly. We multiply $U_c$ and $U_r$ as final uncertainty for each object and high uncertainty means high contribution to network's improvement. And we define a single image as the smallest unit for active learning selection. Then we denote single image's uncertainty as sum of its objects' uncertainty, $U_S$:

$$U_S = \sum (U_C * U_R)$$

We rank all unlabeled images by their uncertainty and select images with more useful information for network's train. In the active learning implementation procedure, as shown in Algorithm[3], we need a few iterations for training the detector with labeled data and annotating images with high uncertainty. Under assistance from active learning, we can train object detector to achieve high mAP with limited annotation cost.

| **Algorithm[3]**: WCR Active Learning Implement Details |
|---|
| **Data:** labeled dataset *I*, unannotated image pool *U*, objects' categories *C*, object detector *D*, iteration times *T*. |
| **Initialization:** Randomly select *I* and train *D* with *I*. |
| **for** *k < T* **do** |
|     **t**est U with D and get bounding boxes *B* with corresponding probability *P* after non-maximal suppression |
|     **for** *each category c in C* **do** |
|         use *I*'s objects' information to calculate $\boldsymbol{W_i^1}$ and $\boldsymbol{W_i^2}$ |
|     **for** *each image i in U* **do** |
|         we calculate each object's WCR uncertainty with $\boldsymbol{W_i^1}$ and $\boldsymbol{W_i^2}$ |
|     we rank *U* with WCR uncertainty and select high ranking images *S* |
|     after annotation by oracle, *I = I + S*, *U = U – S*, |
|     *k = k +* 1 |
| finally, we use *I* to train object detector |

## IV. EXPERIMENT

We experiment on a publicly available remote sensing dataset, DOTA 1.5 which has high quality annotation for 15 categories. It has 1411 images in train dataset and 458 images in validation dataset. Also, it has horizontal and oriented object detection modules and we select the after one in this paper. Because oriented object detection can describe remote sensing objects more carefully than the horizontal one. Moreover, object size differs among all categories and both extremely large and small objects exist in the dataset. Therefore, it's important to select images with both high classification and regression uncertainty for annotation. We implement active learning algorithm on training dataset and evaluate results on validation dataset. We use CenterNet architecture with additional orientation layer and deformable convolution ResNet-18 as backbone for our experiment which proves WCR's capacity. Besides, we experiment some ablation experiments to analyze each part's effect. We implement this experiment with two RTX 2080Ti and below are experiment's details.

*A. Experiment on WCR*

We train CenterNet on a fix part of train dataset and use WCR to select more images for annotation in this section. Firstly, the images in the dataset are usually in size which means they require large computation and memory cost during training and network is hard to converge. Therefore, we crop train images into 1024×1024 pieces and the stride is 824. After that we have 14347 train images while validation images are stationary. We randomly select 10 percent train images for initialization and they keep same in other experiments. This experiment has four iterations. In each iteration, we firstly train detector with annotated images and then WCR selects extra 10 percent cropped images for annotation according to detector's inference on unlabeled data. WCR selects images that have objects owning high classification uncertainty and prevailing bounding box size. WCR aims to achieve high performance on majority of training dataset and regards rare images as outliers. Besides, we implement random selection experiment as comparison. The result is shown in Figure 1 and we can see that our method is very efficient. We can achieve same-level performance with only half annotation amount of random selection. Furthermore, we even achieve higher performance than whole annotation with only half images annotated. For categories with more objects, we select high quality images belonging to majority and few outliers participating the detector's train. Without outliers' influence, detector can achieve higher performance and we implement an ablation experiment in an after part. While the categories with less objects can have better performance with more weights in detector's learning procedure.

*B. Experiment on WC*

In this part, we use classification uncertainty only to select images in experiment iterations. Firstly, we do an ablation experiment on weights which help to overcome imbalances among categories and images. The control group only uses Least Confident (LC) algorithm to calculate each object's uncertainty and the image's uncertainty is sum of objects' uncertainty. Both two experiments have four iterations with same initial dataset. We can see that WC has higher performance and they have different increasement trend. The result is shown in Figure 2 and Table 1. WC sees a drop at the last step because of HC category and we will explain it later. WC shows strong advantage on BD, Bridge, BC and SBF which all have less objects than other categories. The two weights enhance minor categories' selected probabilities and shorten imbalance between all categories. The two weights can partly overcome the remote sensing dataset's class-imbalance and difference in images' objects amount.

Secondly, we design a comparison experiment between WC and WCR to observe their difference and contribution to detector's performance. Without regression uncertainty, the images that have high classification uncertainty and rare bounding box size objects can be selected. In four iterations, an active learning algorithm selects 5800 images. We compare two algorithms' selections and only 729 images are different. The result is shown in Figure 1 and we can see WC has higher performance in early iteration while WCR finally performs a little bit better. They have similar performance on categories except GTF RA and SP. These three categories' objects have a similar size distribution that some objects' sizes are intensive and others' sizes are sparsely distributed. Those sparse objects are not outliers and we can see consecutive increasement in random experiment. With limited annotation cost, it's better to select images from majority firstly and WCR can select objects having similar sizes.

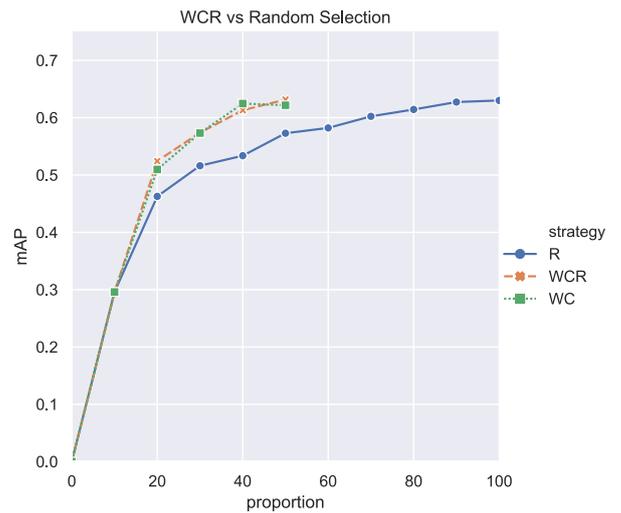

Fig. 1. We compare WC, WCR and random selection's performance in this image. We can see WCR reaching same-level performance with only annotated images.

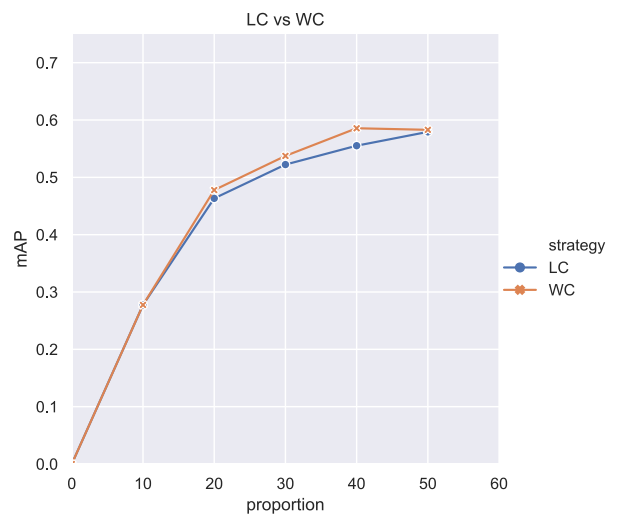

## C. Double weights experiment

With the increasement of annotation images, detector's performance can only be improved a bit with same amount extra annotation as early iteration. Under this condition, it's not worthy to spend more resource on unlabeled data's annotation. Then we turn to dig more from labeled images and we use active learner to evaluate them again. Here we use Least Confident strategy to evaluate all labeled images and the uncertainty is positively related to training difficulty. The images having high uncertainty are hard to learn and deserve higher weights while training. Then we rank them with their uncertainty scores and mark top 20 percent images. Furthermore, we also ranked the unlabeled images with their uncertainty in the experiment of WCR. There are only 10 unlabeled images out of 1450 top ranking images which means we have selected most of helpful images for annotation. This also proves that limited extra annotations are not really helpful to detector's performance and its increasement has met the bottleneck. It's not wise to put large cost in extra images' collection and annotation. We should consider how to gain more from labeled images at this time. To learning marked images better, we should increase their weights while backpropagation. And to achieve that easily, we simply train them twice in each epoch and train the detector again. This measurement doesn't require extra annotation cost and we implement this experiment on both WC and WCR. Our weights adjustment strategy is simple and rigid. It's only used to prove that we can achieve higher performance than fully annotated with no annotation cost. The results are shown in Table 1.

## D. Analysis

Finally, we adopt all methods mentioned above to achieve higher performance with limited annotation cost. Our extra annotated images count around 45 percent of training dataset. They are merged from WC and WCR's selection. Then we use them to train the detector for the first time and evaluate all annotated images' uncertainty with trained detector. Then we rank annotated images according to their uncertainty and double top 20 percent images' train times. The results are shown in Table 1. All experiments have same 10% labeled data for initialization.

From the results we can see that complete annotation doesn't own the highest mAP and we can have the highest mAP with total 55% percent images labeled. Comparing DM and complete annotation's performance, we find that DM has great advantage on categories, SV, LV, Ship, Harbor and HC. HC is a category which only has a small number of objects and is hard to learn. We observe HC's performance in some experiments as shown in Figure 3. They all see an increasing trend and a sharp drop when more images are limited. Even all images are labeled, HC cannot reach the AP peak and DM has the highest AP on HC because some HC images are trained twice. When we apply active learning on remote sensing detection, the categories with few objects deserves more annotation and training weights. While the other four categories are all top 5 ranked by number of objects. More objects do not always bring better performance because there are some outliers decreasing detector's performance. With our active learning algorithm, outliers are not labeled and we can have better performance with less annotation cost. For remote sensing object detection datasets, our active learning algorithm can select helpful images and defense class-imbalance's influence. We even reach higher mAP than complete annotation.

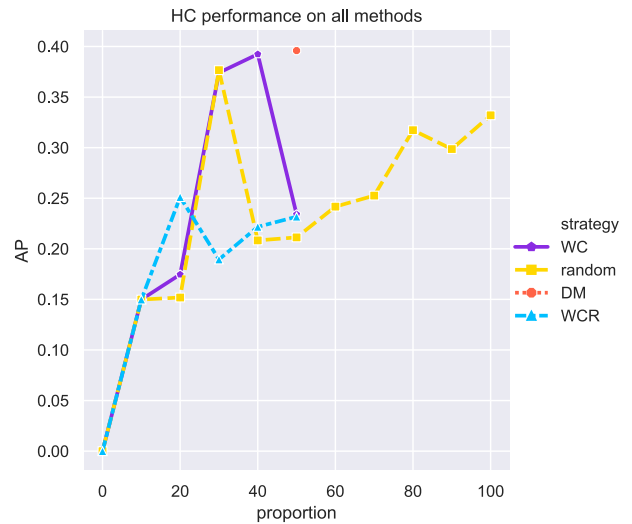

Fig. 3. We display HC's performance under all training methods. They all see a sharp drop with labeled images' amount's increasement.

TABLE I. RESULTS OF ALL EXPERIEMENTS. D DENOTES THE DOUBLE WEIGHTS STRATEGY AND EA DENOTES EXTRA ANNOTATION PERCENT.

| Exp | Plane | BD | Bridge | GTF | SV | LV | Ship | TC | BC | ST | SBF | RA | Harbor | SP | HC | mAP | EA |
|---|---|---|---|---|---|---|---|---|---|---|---|---|---|---|---|---|---|
| LC | 89.5 | 63.5 | 23.1 | 49.3 | 49.0 | 71.7 | 76.8 | 90.7 | 59.6 | 78.8 | 62.3 | 58.0 | 59.7 | 63.8 | 30.2 | 61.78 | 40% |
| WC | 89.6 | 70.0 | 26.0 | 46.9 | 47.6 | 69.5 | 76.6 | 90.6 | 64.7 | 77.8 | 65.5 | 58.8 | 59.3 | 65.7 | 23.4 | 62.17 | 40% |
| WCR | 89.8 | 70.0 | 25.7 | 52.2 | 48.4 | 70.6 | 77.0 | 90.6 | 65.8 | 78.1 | 63.9 | 64.0 | 58.2 | **69.2** | 23.1 | 63.16 | 40% |
| DWC | 89.7 | 70.3 | 28.1 | 49.4 | 47.8 | 69.4 | 76.0 | 90.7 | 64.9 | 78.7 | 61.3 | 61.9 | 59.6 | 65.0 | 23.4 | 62.40 | 40% |
| DWCR | 89.6 | 69.3 | 26.9 | 53.4 | 48.7 | 71.9 | 77.0 | 90.7 | 62.1 | 78.6 | 63.6 | 61.9 | 60.5 | 67.3 | 35.1 | 63.84 | 40% |
| M | 89.8 | 70.0 | 28.0 | **58.3** | 47.6 | 71.2 | 76.4 | 90.6 | 63.2 | 78.6 | 64.3 | 65.2 | 60.4 | 67.7 | 32.5 | 64.00 | 45% |
| DM | 89.7 | 70.8 | 23.9 | 50.4 | **50.3** | **72.4** | **78.3** | 90.6 | 61.9 | **79.2** | 64.2 | 63.2 | **61.5** | 67.4 | **39.5** | **64.26** | 45% |
| R50% | 89.1 | 70.1 | 24.3 | 44.7 | 36.3 | 63.4 | 58.5 | 90.5 | 55.2 | 75.3 | 57.9 | 62.1 | 46.9 | 63.2 | 21.1 | 57.28 | 40% |
| 100% | 89.8 | 72.1 | **28.5** | 54.0 | 45.1 | 66.3 | 69.2 | 90.6 | 60.1 | 78.3 | **66.7** | **66.1** | 57.4 | 66.8 | 33.2 | 63.00 | 100% |

Fig. 2. We compare LC and WC's performance in this image.

CONCLUSION

We propose a novel active learning algorithm for object detection on remote sensing images. It is suitable to any detector which predicts both probabilities for each category and bounding box's width and height. Our method ranks images by their classification and regression uncertainty scores which have two weights to decrease imbalance from the dataset. We especially take objects' regression uncertainty into consider and try hard to select images having major objects but not outliers for annotation. To achieve considerable performance on diverse and large unlabeled dataset, we should use active learning to label important samples and dig more information from labeled samples. In the future work, we may add semi-supervised module to further reduce annotation cost for detector's training.